\documentclass{article}

\usepackage[preprint]{neurips_2026}

\usepackage{subcaption} 
\usepackage[utf8]{inputenc} 
\usepackage[T1]{fontenc}    
\usepackage{hyperref}       
\usepackage{url}            
\usepackage{booktabs}       
\usepackage{amsfonts}       
\usepackage{nicefrac}       
\usepackage{microtype}      
\usepackage{xcolor}         
\usepackage{natbib}
\usepackage{graphicx}
\usepackage[table]{xcolor}
\usepackage{multirow}
\usepackage{siunitx}
\usepackage{multirow}
\usepackage{wrapfig}
\usepackage{hyperref}
\usepackage{ragged2e} 
\hypersetup{
    colorlinks=false, 
    pdfborder={0 0 0}, 
}
\title{Instruction-Evidence Contrastive Dual-Stream Decoding for Grounded Vision-Language Reasoning}

\author{%
  Yashwant Pravinrao Bangde \\
  Department of Computer Science and Engineering\\
  Indian Institute of Technology Kharagpur\\
  Kharagpur, India 721302 \\
  \texttt{yashwantbangde25@kgpian.iitkgp.ac.in} \\
  \AND
  Debaditya Roy \\
  Department of Computer Science and Engineering\\
  Indian Institute of Technology Kharagpur\\
  Kharagpur, India 721302 \\
  \texttt{debaditya@cse.iitkgp.ac.in} \\
}

\begin{document}

\maketitle

\begin{abstract}

  Vision-Language Models (VLMs) exhibit strong performance in instruction following and open-ended vision-language reasoning, yet they frequently generate fluent outputs that are weakly grounded in visual evidence. Prior works have shown that instruction prompting further worsens this issue by amplifying language priors, especially when the visual signal is uncertain or ambiguous. To address this challenge, we propose a decoding framework that explicitly balances linguistic informativeness and visual faithfulness during generation. Our method, Instruction-Evidence Contrastive Dual-Stream Decoding (IECD$^2$), maintains two parallel probability distribution of tokens at each decoding step: an instruction-driven stream that promotes expressive and informative responses, and an evidence-driven stream that enforces strict grounding in the image. These two streams are adaptively fused using a symmetric KL-based contrastive gate, which suppresses tokens favored by language priors but unsupported by visual evidence, while preserving them when both distributions agree. We evaluate IECD$^2$ on multiple datasets spanning various generative vision-language reasoning tasks such as captioning and visual question answering on multiple datasets such as, POPE, MME, VQAv2, AMBER, and MSCOCO. IECD$^2$ demonstrates consistent improvements in task accuracy and reasoning performance with substantial reduction in hallucination compared to state-of-the-art decoding approaches.
\end{abstract}

\section{Introduction}

Vision-language models (VLMs) excel at producing fluent descriptions and answering questions by leveraging rich multimodal representations.
Despite their impressive capabilities, VLMs remain prone to hallucination, producing fluent and plausible descriptions or answers that are not supported by the image. Prior work attributes this behavior to dataset bias and the dominance of language priors when visual evidence is weak or ambiguous (\citet{you-et-al:ferret,zhu-et-al:overcoming-language-priors,zhibo-et-al:overcoming-languague-priors-counterfactual}). To tackle this, Contrastive Decoding (CD) (\citet{li-et-al:cd}) offers a training-free alternative by explicitly contrasting token distributions under different contexts, suppressing tokens driven by spurious or biased language priors while retaining those supported by reliable visual evidence. Building on this idea, Instruction Contrastive Decoding (ICD) (\citet{wang-et-al:instruction-contrastive-decoding}) introduces \emph{instruction disturbance}, where modified instruction prefixes are appended to the original prompt to induce contrasting decoding behaviors. The effectiveness of ICD suggests that decoding behavior in VLMs is highly sensitive to prompting and that different prompts can induce substantially different token distributions. However, ICD treats these behaviors implicitly through prompt perturbations. In contrast, we explicitly construct and reconcile instruction- and evidence-conditioned token distributions during generation.

A consistent trade-off underlies this design -- instruction-aligned prompts induce a distribution that favors informativeness and coverage, whereas prompts that explicitly constrain the model to rely only on visible evidence strongly suppress hallucination but often lead to overly conservative and less descriptive outputs (\citet{liu-et-al:improved_vid}). This reveals a fundamental trade-off between linguistic richness and visual faithfulness. Prompt formulations do not merely control generation style, but actively shift the probability mass toward either language-prior-dominated or evidence-grounded decoding regimes.

Motivated by the complementary yet often conflicting nature of instruction-following and evidence-grounding signals, we propose a dual-stream decoding paradigm that models them as separate but interacting distributions during generation.  \emph{Instruction–Evidence Contrastive Dual-Stream Decoding} (IECD$^2$), instantiates two parallel streams: an instruction stream capturing linguistic priors and an evidence stream enforcing strict visual grounding as shown in Fig.~\ref{fig:Paradigm}.
At each decoding step, both streams independently produce token distributions conditioned on the shared image and question but using different prompts. These distributions are then reconciled through a divergence-aware contrastive gating mechanism, which adaptively modulates token selection based on their agreement. Tokens strongly supported by both streams are promoted, while those favored by the instruction stream but unsupported by evidence are suppressed. This step-wise fusion enables the model to dynamically balance fluency and grounding throughout generation.

Our contributions are as follows: (1) A new contrastive decoding paradigm, IECD$^2$, that contrasts instruction-induced linguistic priors with evidence-constrained distributions. (2) A divergence-aware contrastive gating mechanism that uses disagreement between instruction and evidence streams to suppress ungrounded tokens. (3) IECD$^2$ substantially reduces hallucination and improves grounding accuracy, while maintaining accuracy across diverse captioning and VQA benchmarks such as, MSCOCO, AMBER, VQAv2, LLaVA-Bench and MME.

\begin{figure}[htbp]
    \centering
    \includegraphics[width=\textwidth]{Images/Copy_of_Paradigm.drawio.jpg}
    \caption{Overview of the Instruction–Evidence Contrastive Dual-Stream (IECD$^2$) framework. Text in \textcolor{red}{red} indicates content that is unsupported by the image.}
    \label{fig:Paradigm}
\end{figure}

\section{Related Work}

Hallucination in VLMs refers to the generation of fluent but visually unsupported content, including nonexistent objects, incorrect attributes, or fabricated relations (\citet{liu-et-al:hallucination-vlm,bai-et-al:survey-hallucination}). Earlier work in NLP characterized hallucination as a factual inconsistency with source inputs (\citet{maynez-et-al:hallucination-nlg}), while in vision-language settings, object hallucination has emerged as a central concern (\cite{rohrbach-et-al:chair}). Hallucinations in VLMs are categorized into three types: \emph{category}, \emph{attribute}, and \emph{relation} (\citet{bai-et-al:survey-hallucination}). Most existing methods improve overall visual grounding and factual consistency by mitigating multiple hallucination types simultaneously.

\paragraph{Category Hallucination.}
Category-level hallucination arises when a model predicts the presence of an object class not present in the image, often driven by strong language priors and dataset frequency bias. Inference-time correction methods aim to mitigate this effect without retraining. Post-hoc verification frameworks such as Woodpecker (\citet{yin-et-al:woodpecker}) revise generated outputs using auxiliary detectors or external consistency checks. However, these methods depend on additional models and handcrafted decision rules, introducing error propagation and limiting scalability to open-ended generation. Contrastive decoding (\citet{li-et-al:cd}) offers a training-free alternative by explicitly contrasting token distributions induced by \emph{clean} and \emph{disturbed} inputs, where the disturbance refers to perturbations that weaken visual grounding (e.g., corrupted images or misleading instructions). Visual Contrastive Decoding (VCD) (\citet{leng-et-al:visual-contrastive-decoding}) suppresses category hallucination by contrasting predictions from original and visually perturbed images, thereby down-weighting object tokens that remain confident under visual corruption and are thus likely language-prior driven, while Instruction Contrastive Decoding (ICD) (\citet{wang-et-al:instruction-contrastive-decoding}) exploits disturbance in the instruction space to expose category-level hallucinations amplified by prompt-induced language bias. Multi-Frequency Contrastive Decoding (MFCD) (\citet{liu-et-al:mfcd}) introduces frequency-domain disturbances by selectively removing high- and low-frequency components from images to isolate hallucination-prone distributions. Object-aligned visual contrastive decoding (\citet{chen-et-al:oa-vcd}) masks salient object regions to produce semantically meaningful counterfactual inputs that enhance visual grounding and suppress language-driven object predictions.

\paragraph{Attribute Hallucination.}
Attribute hallucination occurs when a model assigns incorrect properties (e.g., color, size, state) to visually present objects. VACoDe (\citet{kim-et-al:vacode}) addresses this by selecting among multiple image augmentations to maximize contrast, encouraging the model to discount attribute predictions that are unstable under visual variation. Re-Balancing Contrastive Decoding (RBD) (\citet{liang-et-al:rbd}) further targets attribute errors by correcting imbalanced cross-modal attention, reweighting textual and visual contributions to prevent over-dominance of language priors in attribute realization. From an internal representation perspective, IFCD (\citet{wang-et-al:ifcd}) contrasts disturbed hidden states to suppress spurious attribute logits, while VaLiD (\citet{wang-et-al:valid}) introduces uncertainty-aware layer-wise fusion to counteract attribute distortions caused by noisy visual encoding.

\paragraph{Relation Hallucination.}
Relation hallucination refers to incorrect spatial or semantic relationships between objects (e.g., left-right, holding, riding), which often emerge from long-range dependency errors and misaligned multimodal reasoning. Token-level and adaptive strategies such as CATCH (\citet{kan-et-al:catch}) dynamically gate contrastive strength at each decoding step, allowing the model to suppress relation tokens when visual support is weak. Octopus (\citet{suo-et-al:octopus}) further observes that hallucination types vary across decoding stages and proposes a multi-strategy contrastive framework that adaptively switches among disturbance sources to correct relation-level inconsistencies. Extending contrastive decoding to multi-source reasoning, Multi-Path Information Contrastive Decoding (MPI-CD) (\citet{ruan-et-al:mpicd}) introduces a tri-branch framework that contrasts predictions from original, salient-region, and non-salient-region inputs to enhance multimodal reasoning and reduce hallucinations arising from incorrect relational associations.

\section{Methodology}
\label{sec:method}
We introduce Instruction-Evidence Contrastive Dual-Stream Decoding (IECD$^2$), that maintains two token-probability streams during generation: an instruction-guided stream and a visually grounded evidence stream. These streams are fused using a divergence-aware gating mechanism that suppresses instruction-only predictions when they diverge from visual evidence to reduce overdependence on language prior and improve grounded generation. Section~\ref{sec:token-dist} formalizes the dual-stream formulation, while Section~\ref{sec:fusion} presents the divergence-aware fusion mechanism. 

\subsection{Dual-Stream Token Distributions}
\label{sec:token-dist}
Prior work has shown that hallucinations in VLMs are often amplified by instructional prompting, which encourages models to prioritize response fluency and task completion over strict visual grounding (\cite{wang-et-al:instruction-contrastive-decoding}). This motivates the need for methods that can balance both language quality and visual correctness.
To explicitly separate these two behavioral tendencies, IECD$^2$ maintains \emph{two parallel decoding streams} throughout generation, both instantiated from the same underlying VLM
but conditioned via different prompts, shown in Figure \ref{fig:Paradigm}.
At each decoding step $t$ given history $y_{<t}$,
we construct two token probability distributions: Instruction stream $p_t^{(I)}(v)$ and Evidence stream $p_t^{(E)}(v)$ where $v$ is the image.

\paragraph{Instruction stream} models the behavior of a standard instruction-following VLM.
This stream encourages descriptive, coherent and stylistically appropriate responses:
\begin{equation}
p_t^{(I)}(v)
=
p_{\theta_I}
\left(
v \mid y_{<t}, S
\right),
\end{equation}
where $S$ denotes the instruction prompt.
Since typical instructions tend to reward fluency, helpfulness,
and semantic plausibility,
this branch implicitly preserves linguistic prior knowledge.
However, the same property also means that
$p_t^{(I)}$ may assign high probability to visually unsupported objects,
thereby acting as a potential source of hallucination.

\paragraph{Evidence stream.}
To counterbalance the instruction stream, we introduce a second branch that
explicitly constrains generation to rely on visual evidence:
\begin{equation}
p_t^{(E)}(v)
=
p_{\theta_E}
\left(
v \mid y_{<t}, x_{E}
\right),
\end{equation}
where $x_{E}$ is a prompt variant instructing the model
to answer strictly based on observable image content.
Unlike the instruction stream, this branch suppresses unsupported inference
by penalizing uncertain or speculative continuations.
As a consequence,
$p_t^{(E)}$ tends to be conservative, meaning that it assigns high probability only to tokens that are strongly supported by visible evidence and avoids committing to uncertain or weakly grounded entities, but visually faithful.
Notably, no retraining or parameter modification is required and the two token distributions arise due to different conditioning prompts, present in section \ref{appendix:prompts}.
Consequently, IECD$^2$ is fully model-agnostic and applicable to any instruction-tuned VLM.

\begin{figure}[htbp]
    \centering
    \includegraphics[width=\textwidth]{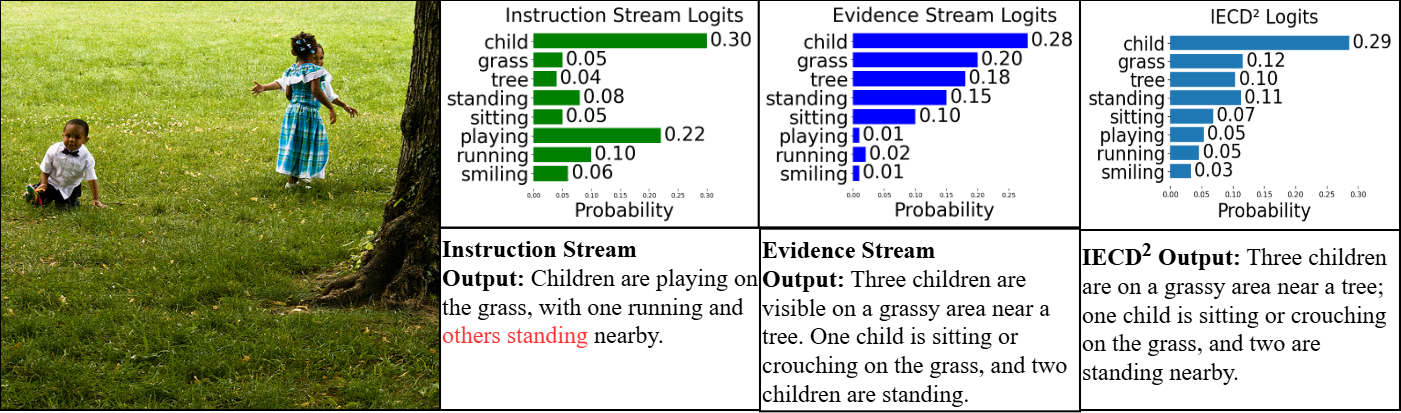}
    \caption{Token probability distributions of the Instruction stream, Evidence stream and IECD$^2$, with their image caption. Text in \textcolor{red}{red} indicates content that is not grounded in the image (i.e., hallucinated or unsupported by the image). IECD$^2$ retains the visual grounding of evidence stream. Image from AMBER (\citet{wang-etal:amber}) dataset.}
    \label{fig:EvidenceStream}
\end{figure}

\subsection{Contrastive Gating via Stream Divergence}
\label{sec:fusion}
Robust multimodal generation requires both linguistic fluency and visual grounding (\citet{li-et-al:cd}). IECD$^2$ achieves this by combining the instruction and evidence token distributions through a gated geometric fusion mechanism building on geometric-mean fusion in  \cite{wang-et-al:instruction-contrastive-decoding}.
Formally, the decoding distribution at step $t$ is defined as
\begin{equation}
p_t^{(\mathrm{IECD^2})}(v)
\propto
\left(p_t^{(I)}(v)\right)^{g_t}
\left(p_t^{(E)}(v)\right)^{1-g_t},
\label{eq:icds_fusion}
\end{equation}
where $g_t\in[0,1]$ controls the relative contribution of the instruction and evidence streams. Converting to log space, we obtain
\begin{equation}
\log p_t^{(\mathrm{IECD^2})}(v)
=
g_t \log p_t^{(I)}(v)
+
(1-g_t)\log p_t^{(E)}(v)
-\log Z_t .
\end{equation}

To adaptively determine $g_t$, we measure the disagreement between the two streams using the symmetric KL divergence (\citet{Jeffreys:skl}):
\begin{equation}
D_t
=
\mathrm{KL}\!\left(p_t^{(I)} \,\middle\|\, p_t^{(E)}\right)
+
\mathrm{KL}\!\left(p_t^{(E)} \,\middle\|\, p_t^{(I)}\right).
\end{equation}

Now, the divergence is assigned to a gating coefficient via
\begin{equation}
g_t
=
\frac{\exp(\eta D_t)}{1+\exp(\eta D_t)},
\end{equation}
where $\eta < 0$ is a contrast-temperature parameter controlling the sensitivity of the gate.

The gating mechanism exhibits different behavior depending upon the value of $D_t$: 1) \textbf{Small divergence ($D_t \approx 0$):}
Both streams agree closely, indicating that the instruction-driven continuation is already consistent with visible content.
In this case, $g_t$ remains moderate, allowing the instruction stream to retain stronger influence, thereby preserving linguistic fluency and descriptive richness. 2) \textbf{Large divergence ($D_t \gg 0$):}
The instruction stream may diverge from the evidence stream when the model relies excessively on language priors, such as predicting commonly expected objects that are not actually present in the image. In such situations, $g_t$ decreases sharply towards 0, reducing the contribution of the instruction stream and encouraging the decoder to rely more on visually grounded evidence for token generation.

The evidence stream alone is insufficient, although it is strictly grounded in visible cues and avoids hallucination, it is overly conservative and often misses higher-level semantics such as actions and interactions. As illustrated in the Fig.~\ref{fig:EvidenceStream}, it assigns low probability to activities like playing or running despite their plausibility, since it focuses on local visual attributes rather than integrating broader contextual and relational information. Consequently, its predictions can be incomplete or overly literal, motivating the need to combine it with the instruction stream in IECD$^2$ to achieve both visual accuracy and semantic completeness. Unlike static weighting strategies (\citet{li-et-al:cd,leng-et-al:visual-contrastive-decoding,wang-et-al:instruction-contrastive-decoding}), contrastive gating enables IECD$^2$ to operate adaptively. 
IECD$^2$ preserves descriptive expressiveness when the model is visually consistent while actively constraining object speculation only when hallucination risk is detected.

\section{Experiments}

\subsection{Experimental Settings}
\paragraph{Benchmarks.} We evaluate IECD$^2$ on benchmarks designed to probe failure modes of vision-language models by studying their impact on hallucination, grounding, and reasoning. To assess hallucination, we use \textbf{POPE} (\cite{li-et-al:pope}) and \textbf{CHAIR}(\cite{rohrbach-et-al:chair}) on the \textbf{MSCOCO} (\cite{lin-et-al:mscoco}) validation set, where POPE measures object hallucination under adversarial and distributional settings and CHAIR evaluates hallucination in generated captions. 
To evaluate grounding fidelity, we use \textbf{AMBER} (\cite{wang-etal:amber}), which includes both caption generation and verification tasks, enabling us to examine whether reductions in hallucination correspond to improved alignment with visual evidence rather than overly conservative predictions. To assess general multimodal reasoning, we evaluate on \textbf{VQAv2} (\cite{goyal-et-al:vqav2}), \textbf{LLaVA-Bench} (\cite{liu-et-al:llava}), and \textbf{MME} (\cite{fu-et-al:mme}) , which span perception, counting, OCR, commonsense reasoning, and open-ended question answering. 

\paragraph{Implementation Details.}  We evaluate IECD$^2$ on two representative vision–language models, LLaVA-1.5-7B (\citet{liu-et-al:llava}) and InstructBLIP (\citet{dai-et-al:instructblip}) (with Vicuna-7B backbone) that are also used by other decoding approaches. These VLMs differ in visual encoders, multimodal alignment, and instruction-tuning strategies, providing a diverse testbed for evaluating hallucination mitigation. For all VLMs across datasets, IECD$^2$ is applied only at decoding time without parameter updates or retraining, enabling a fair and model-agnostic evaluation. For each model, we compare its default decoding with IECD$^2$ using instruction and evidence prompts from Section~\ref{sec:method}. We empirically obtained the best values for instruction temperature $T_I=1.0$, evidence temperature $T_E=0.9$, and $\eta=-3.0$ with symmetric KL divergence for contrastive gating (see Section~\ref{sec:ablation}). Decoding lengths are limited to 3-16 tokens for VQA and 20-64 tokens for captioning, and all experiments are conducted on an NVIDIA L40 48GB GPU.

\subsection{Results on VQA}

\begin{table*}[htbp]
\caption{POPE, AMBER (Discriminative Task), and VQAv2 results across VLMs. Overall, IECD$^2$ consistently delivers strong performance demonstrating an effective balance between instruction following and visual grounding.}

\centering
\resizebox{\linewidth}{!}{
\begin{tabular}{l l
                c c  c c  c c
                c c
                c}
\hline
\multirow{3}{*}{VLM} & \multirow{3}{*}{Method} 
& \multicolumn{6}{c}{\textbf{POPE (MSCOCO)}} 
& \multicolumn{2}{c}{\textbf{AMBER (Dis.)}}
& \multicolumn{1}{c}{\textbf{VQAv2}} \\
\cline{3-11}
& & \multicolumn{2}{c}{Random} 
  & \multicolumn{2}{c}{Popular} 
  & \multicolumn{2}{c}{Adversarial} 
  & Acc. $\uparrow$ & F1 $\uparrow$
  & Acc. $\uparrow$ \\
& & Acc. & F1 & Acc. & F1 & Acc. & F1 
  &  &
  &  \\
\hline

\multirow{11}{*}{\begin{tabular}[c]{@{}c@{}}LLaVA-\\ 1.5\end{tabular}}
 & Instruction prompt only 
           & 83.29 & 81.33 & 81.88 & 80.06 & 78.96 & 77.57 & 69.2 & 72.0
           & 67.54 \\
 & ICD (\cite{wang-et-al:instruction-contrastive-decoding})     
           & 84.23 & 82.06 & 82.73 & 80.70 & 80.23 & 78.52 & 69.9 & 72.8
           & -- \\
 & VCD (\cite{leng-et-al:visual-contrastive-decoding})     
           & 87.73 & 87.16 & 85.38 & 85.06 & 80.88 & 81.33 & 67.3 & 71.1
           & 71.29 \\
 & IFCD (\cite{wang-et-al:ifcd}) 
           & 89.17 & 88.47 & \textbf{88.10} & \textbf{87.36} & \textbf{85.17} & \textbf{84.84} & -- & --
           & -- \\
 & VACoDe (\cite{kim-et-al:vacode})
           & -- & -- & -- & -- & -- & -- & -- & --
           & 72.53 \\
 & RBD (\cite{liang-et-al:rbd})
           & -- & -- & -- & -- & -- & -- & -- & --
           & \underline{78.40} \\
 & Octopus (\cite{suo-et-al:octopus}) 
           & 87.51 & 85.40 & 84.19 & 82.95 & 82.22 & 81.44 & \underline{76.7} & \textbf{82.7}
           & -- \\
 & CATCH (\cite{kan-et-al:catch})   
           & \textbf{90.43} & \textbf{90.13} & 87.07 & 86.56 & 83.17 & 83.18 & -- & --
           & -- \\
 & VaLiD (\cite{wang-et-al:valid})   
           & 89.03 & 88.36 & \underline{87.17} & \underline{86.65} & 83.20 & \underline{83.29} & -- & --
           & -- \\
 & MFCD (\cite{liu-et-al:mfcd})
           & 87.07 & 87.73 & 83.07 & 84.17 & 77.03 & 79.38 & -- & --
           & -- \\
 & Object-aligned VCD (\cite{chen-et-al:oa-vcd})
           & \underline{89.50} & \underline{88.50} & 85.70 & 85.10 & 81.90 & 82.00 & -- & --
           & -- \\
\cline{2-11}
 & \textbf{IECD$^2$ (Ours)} 
           & 88.07 & 86.96
           & 86.70 & 85.68
           & \underline{84.10} & 83.03
           & \textbf{77.8} & \underline{82.6}
           & \textbf{75.30} \\
\hline

\multirow{9}{*}{\begin{tabular}[c]{@{}c@{}}Instruct-\\ BLIP\end{tabular}}
 & Instruction prompt only 
           & 80.71 & 80.41 & 78.22 & 77.33 & 75.84 & 76.59 & 68.2 & 74.6
           & 61.82 \\
 & ICD (\cite{wang-et-al:instruction-contrastive-decoding})     
           & 83.50 & 82.52 & 80.27 & 79.56 & 79.56 & 79.52 & 69.6 & 75.9
           & -- \\
 & VCD (\cite{leng-et-al:visual-contrastive-decoding})     
           & 84.53 & 83.68 & 81.47 & 81.07 & 79.56 & 79.52 & 66.3 & 75.6
           & 66.64 \\
 & IFCD (\cite{wang-et-al:ifcd})    
           & 85.56 & 83.75 & 83.27 & 81.34 & 82.23 & 80.44 & -- & --
           & -- \\
 & VACoDe (\cite{kim-et-al:vacode})
           & -- & -- & -- & -- & -- & -- & -- & --
           & \underline{67.97} \\
 & Octopus (\cite{suo-et-al:octopus}) 
           & 86.63 & 85.30 & \underline{84.90} & 83.55 & \underline{82.83} & 81.43 & \underline{74.0} & \underline{79.7}
           & -- \\
 & CATCH (\cite{kan-et-al:catch})   
           & \textbf{90.17} & \textbf{89.91} & 83.70 & \underline{84.32} & 79.90 & \underline{81.37} & -- & --
           & -- \\
 & VaLiD (\cite{wang-et-al:valid})   
           & 85.90 & 82.98 & 83.43 & 81.59 & 81.33 & 79.59 & -- & --
           & -- \\
\cline{2-11}
 & \textbf{IECD$^2$ (Ours)} 
           & \underline{89.00} & \underline{88.11}
           & \textbf{85.70} & \textbf{85.08}
           & \textbf{83.23} & \textbf{82.94}
           & \textbf{76.6} & \textbf{82.1}
           & \textbf{76.06} \\
\hline
\end{tabular}
}
\label{tab:table1_no_mme}
\end{table*}

\paragraph{POPE (MSCOCO).}
POPE evaluates hallucination under Random (low bias), Popular (frequency bias), and Adversarial (misleading priors) settings. Table~\ref{tab:table1_no_mme} shows that IECD$^2$ improves performance across both LLaVA-1.5 and InstructBLIP, with the largest gains in the Adversarial setting (around 5–6\%), where strong language priors conflict with visual evidence, demonstrating effective suppression of spurious object predictions. In the Popular setting, gains, particularly for InstructBLIP, indicate reduced reliance on frequent object priors, while improvements for LLaVA-1.5 are comparatively smaller, likely because LLaVA-1.5 already achieves stronger multimodal perception and instruction-following performance across a broad range of benchmarks~\cite{liu-et-al:improved_vid}. In the Random setting, improvements are more modest (around 4–5\%) since hallucination pressure is lower and predictions are already largely grounded. Across all POPE settings, IECD$^2$ is particularly effective in mitigating object-level hallucinations caused by dataset frequency and co-occurrence biases, as the evidence stream counteracts misleading high-probability object tokens that are favored by the instruction stream under adversarial and popular priors.

\paragraph{AMBER Discriminative Task.}
The AMBER Discriminative Task evaluates visual grounding by measuring how well predictions align with image evidence, despite the benchmark containing visually ambiguous and semantically correlated object categories that make grounding particularly challenging. As shown in Table~\ref{tab:table1_no_mme}, IECD$^2$ consistently improves grounding performance across VLMs, achieving 77.8/82.6 on LLaVA-1.5 compared to 69.2/72.0 with direct decoding, and 76.6/82.1 on InstructBLIP compared to 68.2/74.6 with direct decoding. The significance of these gains lies in the fact that discriminative settings often contain subtle visual distinctions where models can answer correctly using dataset priors or semantic associations without true visual verification (\cite{wang-etal:amber}). IECD$^2$ reduces this tendency by explicitly contrasting instruction-driven predictions against evidence-grounded predictions, leading to larger improvements when language priors conflict with weak or ambiguous visual cues. Overall, the results indicate that IECD$^2$ improves fine-grained visual verification by reducing reliance on semantic similarities and enforcing stronger consistency between generated predictions and image evidence.

\paragraph{VQAv2.}
VQAv2 measures general visual question answering capability under diverse linguistic and visual conditions. Table~\ref{tab:table1_no_mme} shows that IECD$^2$ improves answer reliability while maintaining competitive task . For LLaVA-1.5, IECD$^2$ achieves 75.30\%, outperforming Instruction prompt only decoding (67.54\%) and VCD (71.29\%), and approaching the 78.4\% reported by RBD. For InstructBLIP, it attains 76.06\%, improving over Instruction prompt only (61.82\%) and VCD (66.64\%), while also surpassing VACoDe. Improvements are most evident in cases requiring precise visual grounding, where incorrect answers often arise from language priors overriding image evidence, indicating that IECD$^2$ helps by suppressing visually unsupported token choices. Overall, the results suggest that IECD$^2$ is particularly effective for visually grounded question answering because VQAv2 contains diverse open-ended questions where semantically plausible answers may not always align with the actual visual content.

\begin{wrapfigure}{r}{0.4\textwidth}
    \centering
    \vspace{-30pt}
    \includegraphics[width=\linewidth]{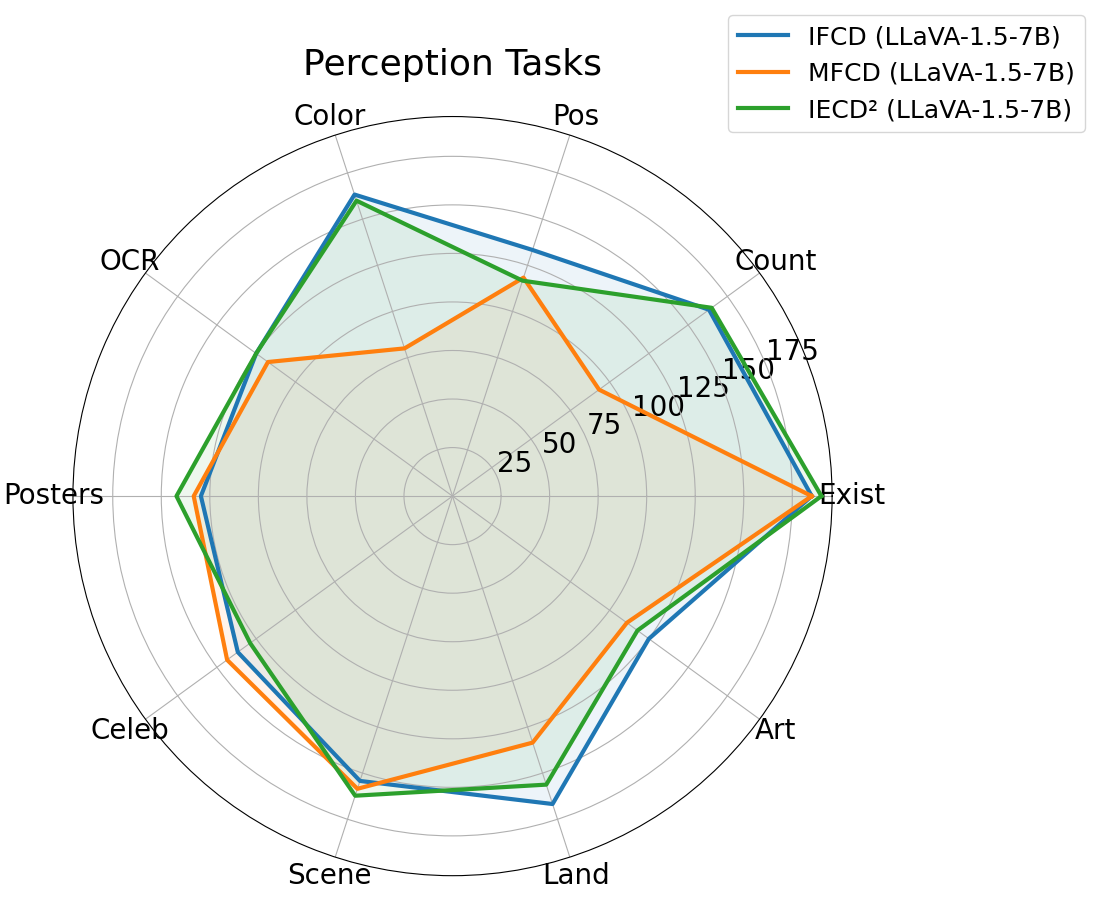}
    \caption{Comparison of Accuracy+ on MME perception tasks involving object, attribute, and relation hallucination.}
    \label{fig:perception}
    \vspace{-10pt}
\end{wrapfigure}
\paragraph{Results on Perception Tasks.} MME Perception tests visual grounding and compositional question answering across 8 perception tasks. As shown in Fig.~\ref{fig:perception}, IECD$^2$ performs particularly well on object-centric and contextual tasks such as Existence, Counting, Scene understanding, and Landmark recognition, suggesting that jointly modeling instruction semantics and visual evidence effectively suppresses hallucinated object predictions and improves consistency with image content. Compared to MFCD (\cite{liu-et-al:mfcd}) and IFCD (\cite{wang-et-al:ifcd}), IECD$^2$ shows stronger robustness in tasks such as Existence, Counting, Color recognition, Scene understanding, and Landmark recognition, where language priors can conflict with global visual context. IECD$^2$ does not perform as well on fine-grained localization tasks such as Position understanding and OCR, which depend heavily on precise spatial alignment and localized textual cues. As IECD$^2$ primarily operates through token-level contrastive interactions between instruction and evidence streams, it is more effective at mitigating high-level semantic hallucinations than recovering subtle localized details. 

\subsection{Results on Captioning}

\paragraph{MSCOCO Captioning.}
To quantify object-level hallucination in caption generation, we report CHAIR\textsubscript{s} and CHAIR\textsubscript{i}. Since prior methods report results on an unspecified random subset of 500 images that is not publicly available, we evaluate our method on 5 independently sampled subsets of 500 images and report the average performance. As shown in Table~\ref{tab:table2}, IECD$^2$ achieves the strongest hallucination suppression across both VLMs, reducing CHAIR scores by over \textbf{85\%} on LLaVA-1.5 and up to \textbf{98\%} on InstructBLIP relative to direct decoding, while consistently outperforming strong baselines such as IFCD and CATCH. The gains are consistent across both CHAIR\textsubscript{s} and CHAIR\textsubscript{i}, indicating fewer hallucinated object mentions and substantially improved instance-level grounding. Overall, the results suggest that IECD$^2$ is particularly effective at mitigating object hallucinations in open-ended caption generation, where language priors frequently introduce plausible but visually unsupported objects.

\paragraph{AMBER Generative Task.}
IECD$^2$ consistently achieves the lowest hallucination across both models while maintaining competitive coverage. As shown in Table~\ref{tab:table2}, for LLaVA-1.5, hallucination (Hal) reduces from 44.5 to \textbf{5.6} and cognitive hallucination (Cog) from 31.0 to \textbf{0.4}. For InstructBLIP, Hal decreases from 31.8 to \textbf{10.4} and Cog from 2.6 to \textbf{0.5}. These correspond to substantial reductions, reaching up to \textbf{87--98\%} for LLaVA-1.5 and strong improvements for InstructBLIP. While coverage decreases compared to Instruction prompt only decoding, it remains comparable to or better than several hallucination mitigation baselines. The relatively larger coverage reduction for InstructBLIP suggests that the evidence-guided contrastive decoding suppresses a greater number of uncertain object predictions, indicating stronger reliance on semantically plausible but weakly grounded tokens during generation. In AMBER generative task, reducing hallucinated objects is more critical than maximizing coverage because unsupported object mentions directly degrade grounding exactness and output reliability.

\begin{table}[htbp]
\caption{Hallucination and grounding evaluation across MSCOCO and AMBER (Generative Task). IECD$^2$ consistently produces superior results to prior methods in hallucination reduction while preserving strong task performance across benchmarks.}
\centering
\scriptsize
\setlength{\tabcolsep}{7pt}
\begin{tabular}{p{0.5cm} l c c c c c c}
\hline
\multirow{2}{*}{VLM} & \multirow{2}{*}{Method}
& \multicolumn{2}{c}{\textbf{MSCOCO}}
& \multicolumn{4}{c}{\textbf{AMBER (Gen.)}} \\
\cline{3-8}
&
& CH$_s$ ($\downarrow$) & CH$_i$ ($\downarrow$)
& CH. ($\downarrow$) & Cov. ($\uparrow$) & Hal ($\downarrow$) & Cog ($\downarrow$) \\
\hline

\multirow{9}{*}{\begin{tabular}[c]{@{}c@{}}LLaVA-\\ 1.5\end{tabular}}
 & Instruction prompt only & 20.0 & 15.2 & 9.2 & 38.0 & 44.5 & 31.0 \\
 & ICD (\cite{wang-et-al:instruction-contrastive-decoding})     & 50.8 & 16.9 & 8.2 & \underline{44.5} & 31.0 & 2.2 \\
 & VCD (\cite{leng-et-al:visual-contrastive-decoding})     & 20.5 & 7.00 & 8.2 & 41.5 & 36.3 & 3.1 \\
 & VACoDe (\cite{kim-et-al:vacode})  & --   & --   & --  & --   & --   & --  \\
 & RBD (\cite{liang-et-al:rbd})     & -- & -- & -- & -- & -- & -- \\
 & IFCD (\cite{wang-et-al:ifcd})    & \underline{13.2} & 5.6 & -- & -- & -- & -- \\
 & CATCH (\cite{kan-et-al:catch})   & 14.3 & \underline{4.7} & -- & -- & -- & -- \\
 & MFCD (\cite{liu-et-al:mfcd})    & 54.0 & 15.0 & -- & -- & -- & -- \\
 & Octopus (\cite{suo-et-al:octopus}) & --   & --   & \underline{4.8} & \textbf{49.2} & \underline{23.4} & \underline{1.2} \\
\cline{2-8}
 & \textbf{IECD$^2$ (Ours)} & \textbf{2.96} & \textbf{2.25} & \textbf{2.3} & 40.5 & \textbf{5.6} & \textbf{0.4} \\
\hline
\hline

\multirow{8}{*}{\begin{tabular}[c]{@{}c@{}}Instruct-\\ BLIP\end{tabular}}
 & Instruction prompt only & 48.0 & 13.9 & 8.4 & 44.4 & 31.8 & 2.6 \\
 & ICD (\cite{wang-et-al:instruction-contrastive-decoding})    & 58.2 & 18.5 & 8.1 & 42.4 & 31.0 & 2.2 \\
 & VCD (\cite{leng-et-al:visual-contrastive-decoding}) & 21.7 & 7.7 & 8.6 & \underline{45.0} & 31.8 & 2.6 \\
 & VACoDe (\cite{kim-et-al:vacode})  & -- & -- & -- & -- & -- & -- \\
 & IFCD (\cite{wang-et-al:ifcd})    & 39.6 & 11.2 & -- & -- & -- & -- \\
 & CATCH (\cite{kan-et-al:catch})   & \underline{7.2} & \underline{1.8} & -- & -- & -- & -- \\
 & Octopus (\cite{suo-et-al:octopus}) & --   & --   & \underline{6.1} & \textbf{48.5} & \underline{22.2} & \underline{1.3} \\
\cline{2-8}
 & \textbf{IECD$^2$ (Ours)} & \textbf{0.64} & \textbf{0.85} & \textbf{3.8} & 39.6 & \textbf{10.4} & \textbf{0.5} \\
\hline
\end{tabular}
\label{tab:table2}
\end{table}

\subsection{Ablation Studies}\label{sec:ablation}
\paragraph{Effect of Divergence Measure.} We analyze the effect of different divergence measures used in the contrastive gate of IECD$^2$ on the AMBER Generative benchmark and MSCOCO CHAIR evaluation with LLaVA-1.5-7B, where for MSCOCO we report results on a randomly sampled subset of 500 images. Table~\ref{tab:kl_ablation_combined} presents an ablation over multiple divergence choices, including forward KL (\cite{Kullback-et-al:kldivergence}), reverse KL (\cite{Kullback-et-al:kldivergence}), symmetric KL (\cite{Jeffreys:skl}), Hellinger distance (\cite{Hellinger:HDist}), and Bhattacharyya distance (\cite{Bhattacharyya:bhattacharyadistance}). Among these, symmetric KL consistently yields the best trade-off between hallucination reduction and coverage, achieving the lowest CHAIR, Hal, and Cog scores while significantly improving coverage on AMBER. Forward and reverse KL perform comparably but are slightly less effective, likely due to their directional bias. In contrast, Hellinger and Bhattacharyya distances, which are bounded and less sensitive to fine-grained probability differences, result in weaker contrast signals and marginally higher hallucination.

\vspace{-8pt}

\begin{table}[htbp]
\caption{Effect of different divergence measures on contrastive gates}
\scriptsize
\centering

\begin{subtable}{0.48\linewidth}
\centering
\caption{AMBER Generative Task}
\begin{tabular}{l c c c c}
\hline
\textbf{Divergence} & \textbf{CHAIR} $\downarrow$ & \textbf{Cover} $\uparrow$ & \textbf{Hal} $\downarrow$ & \textbf{Cog} $\downarrow$ \\
\hline
Forward KL   & 2.6 & 34.9 & 5.8 & 0.4 \\
Reverse KL   & 2.6 & 34.8 & 5.9 & 0.4 \\
Hellinger Distance & 2.6 & 34.9 & 5.9 & 0.4 \\
Bhattacharyya Distance & 2.5 & 34.8 & 5.9 & 0.4 \\ \hline
\textbf{Symmetric KL} & \textbf{2.3} & \textbf{40.5} & \textbf{5.6} & \textbf{0.4} \\
\hline
\end{tabular}

\end{subtable}
\hfill
\begin{subtable}{0.48\linewidth}
\centering
\caption{MSCOCO CHAIR}
\begin{tabular}{l c c}
\hline
\textbf{Divergence} & \textbf{CHAIR\textsubscript{s}} $\downarrow$ & \textbf{CHAIR\textsubscript{i}} $\downarrow$ \\
\hline
Forward KL   & 5.40 & 2.75 \\
Reverse KL   & 5.20 & 2.70 \\
Hellinger Distance & 4.80 & 2.53 \\
Bhattacharyya Distance & 5.80 & 2.86 \\ \hline
\textbf{Symmetric KL} & \textbf{4.71} & \textbf{2.48} \\
\hline
\end{tabular}
\end{subtable}

\label{tab:kl_ablation_combined}
\end{table}
\begin{wraptable}{r}{0.52\linewidth}
\vspace{-18pt}
\caption{Effect of IECD$^2$ hyperparameters on the AMBER generative task.}
\centering
\scriptsize
\setlength{\tabcolsep}{3pt}
\begin{tabular}{l c c c c c}
\hline
\textbf{Parameter} & \textbf{Value}
& \textbf{CHAIR}$\downarrow$
& \textbf{Cov.}$\uparrow$
& \textbf{Hal}$\downarrow$
& \textbf{Cog}$\downarrow$ \\
\hline

\multirow{3}{*}{\shortstack[l]{Gating Strength $\eta$\\
($T_I{=}1.0,\ T_E{=}0.9$)}}
& $-2.0$ & 2.4 & \textbf{41.2} & 5.8 & 0.4 \\
& $-3.0$ & 2.3 & 40.5 & 5.6 & 0.4 \\
& $-5.0$ & \textbf{2.2} & 39.6 & \textbf{5.5} & 0.4 \\

\hline

\multirow{3}{*}{\shortstack[l]{Instr. Temp. $T_I$\\
($T_E{=}0.9,\ \eta{=}-3.0$)}}
& 0.7 & \textbf{2.2} & \textbf{40.9} & 5.4 & 0.4 \\
& 1.0 & 2.3 & 40.5 & 5.6 & 0.4 \\
& 1.3 & \textbf{2.2} & 38.6 & \textbf{5.0} & 0.4 \\

\hline

\multirow{3}{*}{\shortstack[l]{Evid. Temp. $T_E$\\
($T_I{=}1.0,\ \eta{=}-3.0$)}}
& 0.7 & \textbf{2.2} & 39.1 & \textbf{5.2} & 0.4 \\
& 0.9 & 2.3 & 40.5 & 5.6 & 0.4 \\
& 1.3 & 2.3 & \textbf{41.5} & 5.6 & 0.4 \\

\hline
\end{tabular}
\label{tab:amber_ablation_tempeta}
\vspace{-10pt}
\end{wraptable}
\paragraph{Effect of Temp. and Gating Strength}\label{sec:ablation_temp_eta} The instruction and evidence temperatures control the sharpness of their respective token distributions: lower values produce more confident predictions, while higher values yield more diverse outputs (\cite{Hinton-et-al:distillingknowledgeneuralnetwork}). To study the sensitivity of IECD$^2$ to decoding hyperparameters, we perform ablations on the AMBER Generative Task using LLaVA-1.5-7B by varying the instruction temperature $T_I$, evidence temperature $T_E$, and contrastive gating strength $\eta$, reporting CHAIR, Coverage, Hallucination rate (Hal), and Cognitive Hallucination (Cog) in Table~\ref{tab:amber_ablation_tempeta}. Stronger contrastive gating (more negative $\eta$) consistently reduces hallucination while maintaining stable Coverage, indicating that the KL-guided gate suppresses unsupported tokens without degrading descriptive completeness. Lower $T_E$ further improves visual grounding, whereas higher $T_I$ increases linguistic diversity with only minor changes in Coverage. Overall, the ablation results show that hallucination reduction in IECD$^2$ is primarily driven by stronger evidence-guided contrastive suppression, while the temperature parameters provide controllable trade-offs between grounding strength and generation diversity.

\begin{figure*}[!htbp]
\centering
\begin{subfigure}{0.495\textwidth}
    \centering
    \includegraphics[width=\linewidth]{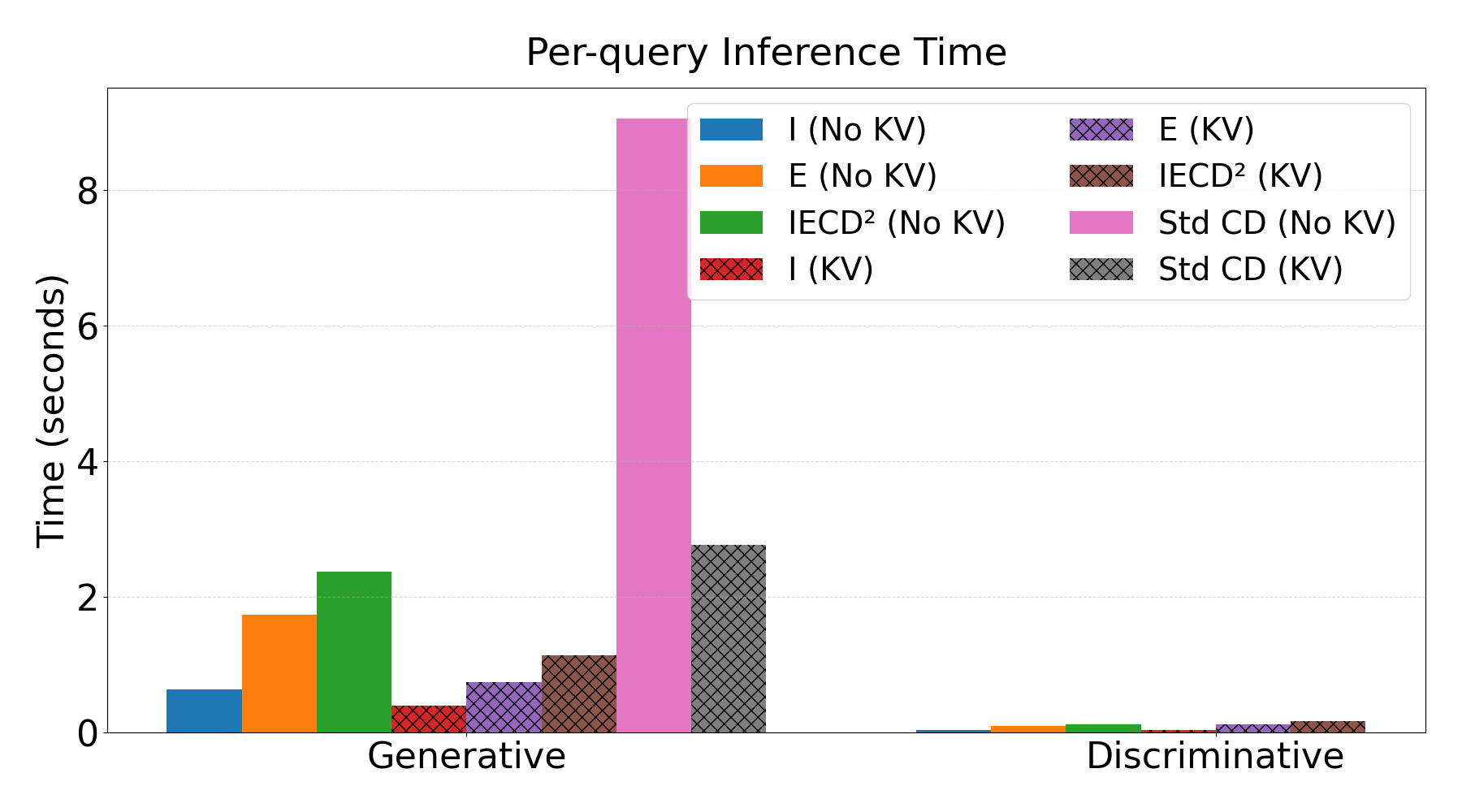}
\end{subfigure}
\hfill
\begin{subfigure}{0.495\textwidth}
    \centering
    \includegraphics[width=\linewidth]{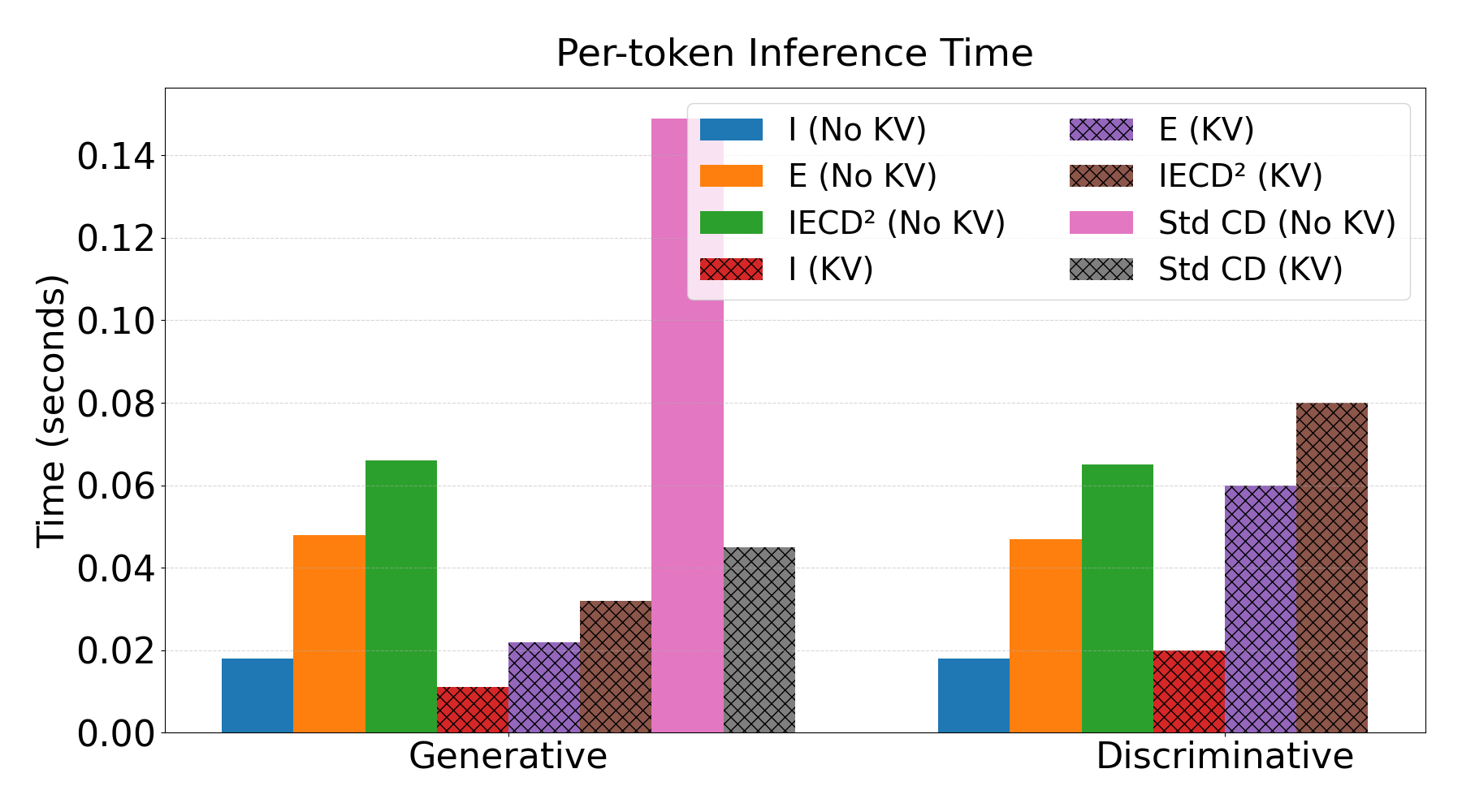}
\end{subfigure}

\caption{Comparison of Runtime analysis of IECD$^2$ with Instruction-only and Evidence-only prompts (with and without KV caching).}
\label{fig:runtime_analysis}
\end{figure*}

\paragraph{Inference Time.}
IECD$^2$ exhibits linear scaling with sequence length, with runtime equal to the sum of the instruction and evidence streams as shown in Figure~\ref{fig:runtime_analysis}. IECD$^2$ remains comparable to standard contrastive decoding, both with key-value caching and without key-value caching for generative tasks . The fusion of two streams introduces negligible overhead ($<6\%$) while evidence reuse with key-value caching prevents any increase in context length or computational cost. Since discriminative tasks generate only a few tokens, KV-cache reuse provides limited benefit, causing fixed dual-stream decoding overhead to dominate runtime.

\vspace{-8pt}

\begin{figure}[htbp]
    \centering
    \hspace*{-0.7cm}
    \includegraphics[width=\columnwidth]{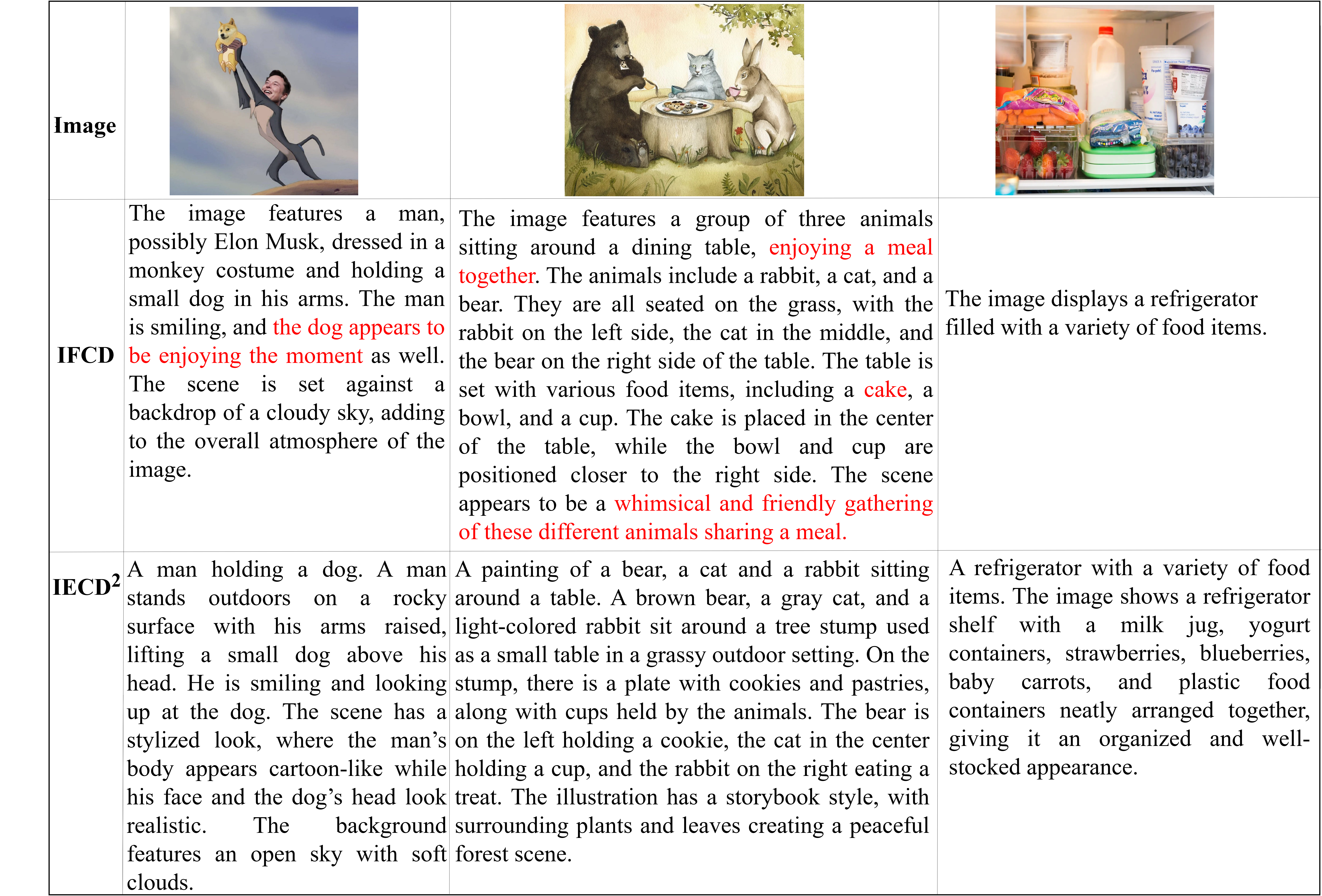}
    \caption{Comparison of created descriptions by IFCD (\cite{wang-et-al:ifcd}) and IECD$^2$ on Llava-Bench (\cite{liu-et-al:llava}) dataset. Text in \textcolor{red}{red} indicates content that is not grounded in the image (i.e., hallucinated or unsupported by the image).}
    \label{fig:Amber_QA}
\end{figure}

\paragraph{Qualitative Analysis.} To analyze the qualitative behavior of IECD$^2$, we visualize captioning results on the LLaVA-Bench (\cite{liu-et-al:llava}) dataset, as shown in Fig.~\ref{fig:Amber_QA}, and compare IECD$^2$ with IFCD (\cite{wang-et-al:ifcd}) to highlight differences in visual grounding and reliance on language priors. In the first example, IFCD speculates about the person’s identity and the dog’s emotions, while IECD$^2$ simply describes a stylized image of a man lifting a small dog against a cloudy sky. In the second example, IFCD assigns emotions and specific food categories to illustrated animals, whereas IECD$^2$ describes three animals seated around a tree-stump table with food items. In the final example, IECD$^2$ lists the visible contents of a refrigerator shelf, while IFCD gives only a vague summary. Overall, IECD$^2$ produces short, accurate, and visually grounded captions, avoiding identity speculation and unsupported semantic inferences that frequently appear in IFCD outputs.

\section{Limitations}\label{sec:limitations}
The approach depends on prompt-based separation of instruction and evidence, and its performance is influenced by the quality of visual grounding available in the input. In cases where visual evidence is insufficient or ambiguous, improvements in grounding may be limited. Furthermore, the divergence-based gating mechanism introduces additional hyperparameters whose behavior may vary across tasks and model architectures.

\section{Conclusion}
We presented Instruction-Evidence Contrastive Dual-Stream Decoding (IECD$^2$), a decoding-time framework that improves the reliability of vision-language model generation by explicitly contrasting instruction-driven and evidence-grounded token distributions. By maintaining two parallel probability streams and fusing them through a symmetric KL-based adaptive gate, IECD$^2$ dynamically suppresses visually unsupported continuations while preserving linguistic fluency when the two distributions are consistent. This formulation enables the model to balance language priors and visual grounding at each decoding step, rather than committing to a single prompting regime. Extensive experiments across both generative and discriminative vision-language reasoning tasks show that IECD$^2$ consistently reduces hallucination and improves grounding accuracy, demonstrating that contrastive, divergence-aware fusion at inference time is an effective and general mechanism for enhancing the robustness of modern VLM decoding.

{
\small
\bibliographystyle{plainnat}
\bibliography{neurips}


}


\appendix

\section{Technical appendices}

\subsection{Results on Cognition Tasks}
We analyze the behavior of IECD$^2$ on the MME benchmark using LLaVA-1.5 to study how instruction and evidence streams contribute to cognition-oriented reasoning tasks. As shown in Fig.~\ref{fig:perception}, IECD$^2$ performs strongly on commonsense and code reasoning tasks, indicating that the interaction between instruction and evidence streams helps maintain contextual consistency and suppress semantically plausible but unsupported generations. These tasks benefit from balancing instruction semantics with controlled token selection, allowing the model to generate more coherent and reliable reasoning outputs. In contrast, performance is comparatively weaker on numerical calculation and translation tasks, which rely heavily on precise symbolic operations and exact token-level prediction. Because IECD$^2$ primarily improves grounding and semantic consistency rather than explicit symbolic reasoning, its advantages are less pronounced in tasks where correctness depends on strict numerical accuracy or exact linguistic mapping. Overall, the results indicate that IECD$^2$ is particularly effective for contextual and reasoning-intensive cognition tasks, while purely symbolic tasks remain challenging.

\begin{figure}[htbp]
    \centering
    \includegraphics[width=0.5\textwidth]{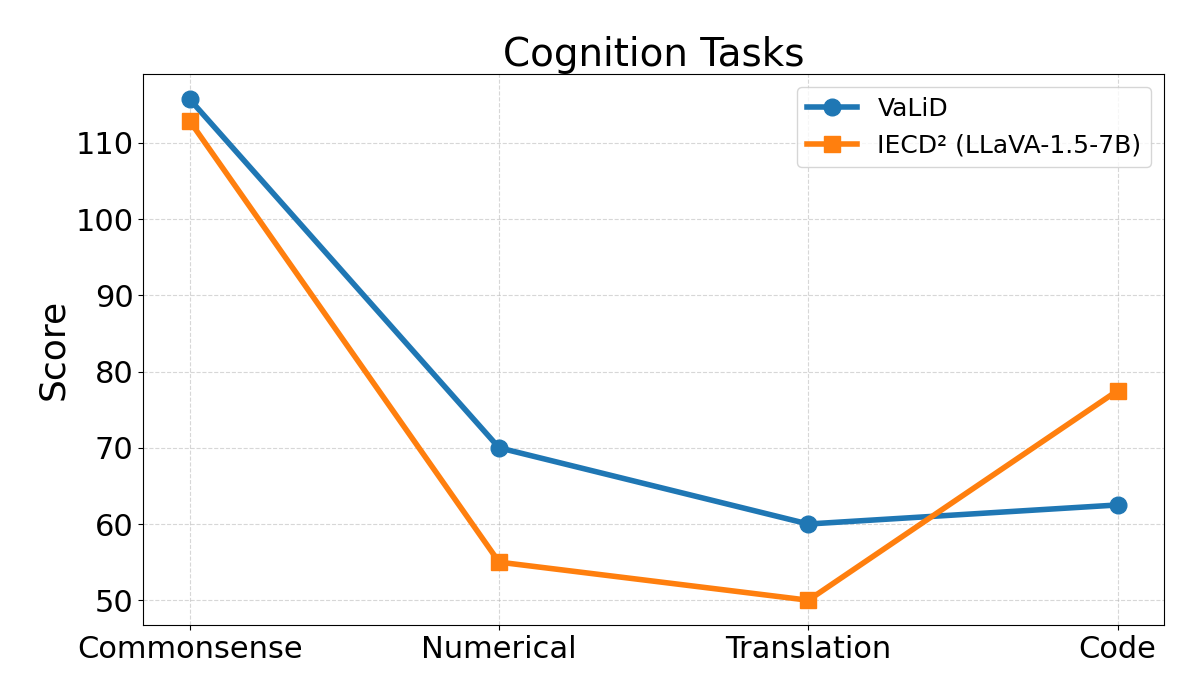}
    \caption{Results on cognition tasks covering diverse hallucination scenarios across commonsense reasoning, numerical calculation, text translation and code reasoning.}
    \label{fig:Cognition}
\end{figure}

\subsection{Effect of Dual-Stream Design.}  
To verify that the dual-stream formulation in IECD$^2$ is necessary for balancing grounding and fluency, we conduct an ablation study on the MSCOCO captioning task using LLaVA-1.5-7B, comparing against \textit{Instruction-only} decoding (standard instruction-aligned generation) and \textit{Evidence-only} decoding (strict visual grounding), with CHAIR\textsubscript{s} and CHAIR\textsubscript{i} reported in Table~\ref{tab:icds_ablation_ds}. MSCOCO is particularly suitable for this analysis because it contains diverse real-world scenes with multiple objects, attributes, and contextual interactions, enabling evaluation of both visual grounding and descriptive generation~(\cite{lin-et-al:mscoco}). In addition, the availability of dense object annotations and the established use of CHAIR metrics on MSCOCO provide reliable measurement of hallucination behavior under different decoding strategies~(\cite{rohrbach-et-al:chair}).
The \textit{Instruction-only} baseline exhibits the highest hallucination among the two individual streams, confirming that instruction-aligned decoding relies heavily on language priors and tends to introduce visually unsupported objects. In contrast, the \textit{Evidence-only} variant substantially reduces hallucination by enforcing strict grounding, but often produces overly conservative and less expressive captions. IECD$^2$ achieves lower hallucination than both individual streams, reducing CHAIR\textsubscript{s} from 4.48 to 2.96 and CHAIR\textsubscript{i} from 3.12 to 2.25 relative to \textit{Instruction-only}, while also outperforming \textit{Evidence-only} decoding. 
Overall, these results demonstrate that the dual-stream formulation is central to the effectiveness of IECD$^2$, rather than a superficial architectural modification.

\centering
\setlength{\tabcolsep}{2pt}
\captionof{table}{Comparing IECD$^2$ dual-stream design versus individual streams and Contrastive Decoding}
\scriptsize
\begin{tabular}{lcc}
\hline
\textbf{Method} & \textbf{CHAIR\textsubscript{s}} $\downarrow$ & \textbf{CHAIR\textsubscript{i}} $\downarrow$ \\
\hline
Instruction only & 4.48 & 3.12 \\
Evidence only    & 3.16 & 2.30 \\
\hline
IECD$^2$  & \textbf{2.96} & \textbf{2.25} \\
\hline
\end{tabular}
\label{tab:icds_ablation_ds}

\subsection{Prompts Used in IECD$^2$}
\label{appendix:prompts}
\normalsize 

This section presents the exact prompts used for the instruction and evidence streams across different task settings in IECD$^2$.

\subsubsection{Image Captioning}

\textbf{Instruction Prompt}

\begin{verbatim}
Describe the image in detail.
Caption:
\end{verbatim}

\textbf{Evidence Prompt}

\begin{verbatim}
Describe ONLY what is clearly visible in the image. Do not guess.
Caption:
\end{verbatim}

\subsubsection{Yes/No Visual Question Answering}

\textbf{Instruction Prompt}

\begin{verbatim}
Answer the question based on the image. Answer only yes or no.

Question: {question}
Answer:
\end{verbatim}

\textbf{Evidence Prompt}

\begin{verbatim}
Answer the question using only visible evidence in the image.
Do not guess. Answer only yes or no.

Question: {question}
Answer:
\end{verbatim}

\subsubsection{Open-ended Visual Question Answering}

\textbf{Instruction Prompt}

\begin{verbatim}
Answer the visual question briefly.

Question: {question}
Answer:
\end{verbatim}

\textbf{Evidence Prompt}

\begin{verbatim}
Answer ONLY using visible information from the image.
Do not assume anything.

Question: {question}
Answer:
\end{verbatim}

\subsection{Additional Qualitative Analysis}
\begin{figure}[htbp]
    \includegraphics[width=\columnwidth]{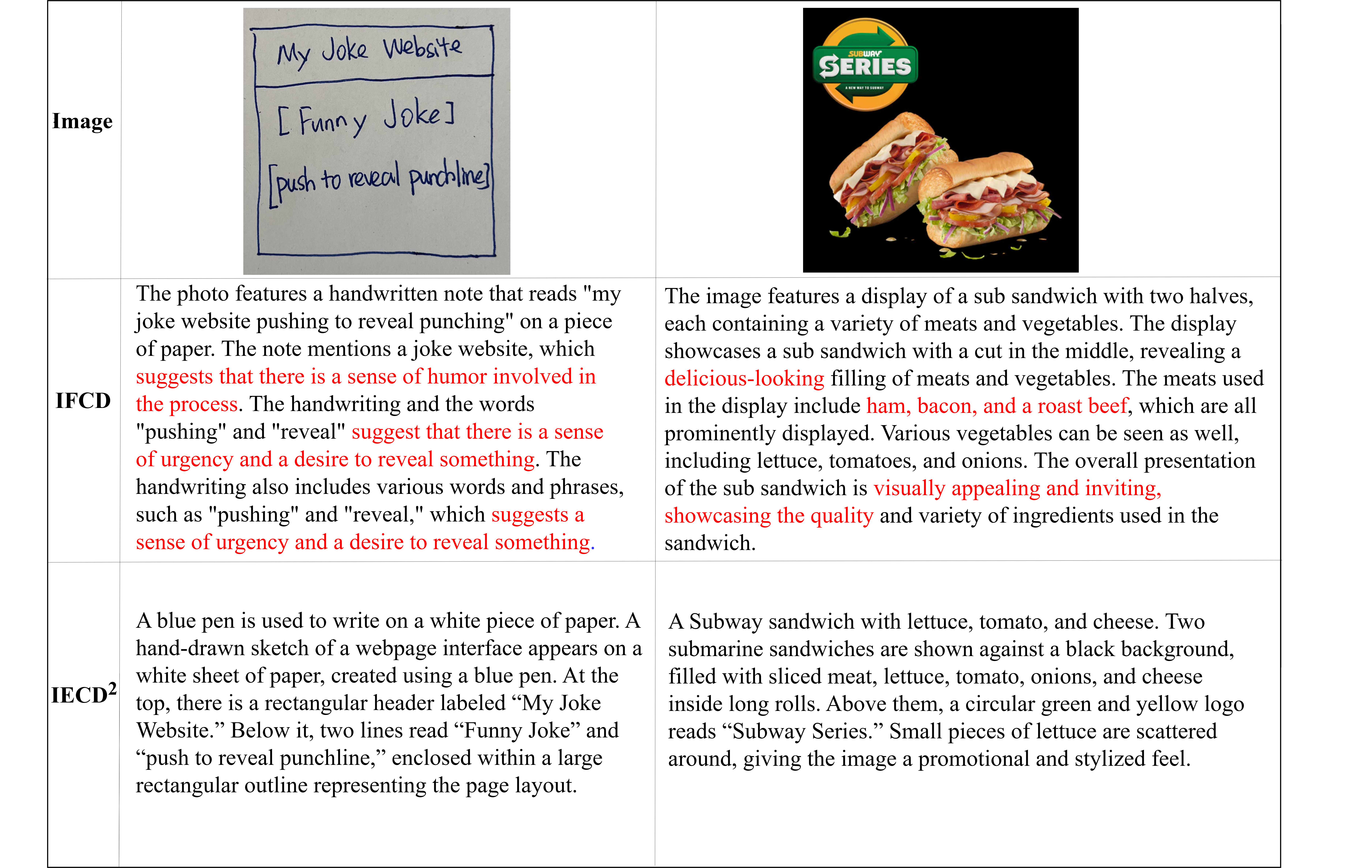}
    \caption{Comparison of created descriptions by IFCD (\cite{wang-et-al:ifcd}) and IECD$^2$ on Llava-Bench (\cite{liu-et-al:llava}) dataset. Text in \textcolor{red}{red} indicates content that is not grounded in the image (i.e., hallucinated or unsupported by the image).}
    \label{fig:Qual_more}
\end{figure}

\justifying
To further analyze the qualitative behavior of IECD$^2$, we present additional captioning examples in the Appendix comparing IECD$^2$ with IFCD (\cite{wang-et-al:ifcd}). As shown in Fig.~\ref{fig:Qual_more}, IFCD often introduces semantic interpretations and subjective inferences that are not directly supported by the image content. In the first example, IFCD infers humor, urgency, and an intention to reveal something from a hand-drawn sketch labeled “My Joke Website,” despite these concepts not being visually observable. In contrast, IECD$^2$ provides a grounded description of the handwritten webpage sketch, explicitly describing the visible text and layout elements. In the second example, IFCD describes the sandwich advertisement using subjective phrases such as “delicious-looking,” “visually appealing,” and references specific meat categories beyond what can be confidently verified from the image. IECD$^2$, however, focuses on observable visual details, describing the sandwiches, vegetables, logo, and presentation without introducing unsupported semantic assumptions. These examples further demonstrate that IECD$^2$ generates more precise and visually grounded captions while reducing speculative reasoning and language-prior-driven hallucinations.

\end{document}